\pdfoutput=1

\documentclass[11pt]{article}

\usepackage{authblk}

\usepackage[]{acl}

\usepackage{times}
\usepackage{latexsym}

\usepackage{graphicx}
\graphicspath{ {./} }
\usepackage{multirow}
\usepackage{url}
\usepackage{amsmath}
\usepackage{booktabs}
\usepackage[T1]{fontenc}

\usepackage[utf8]{inputenc}

\usepackage{microtype}

\title{DPTDR: Deep Prompt Tuning for Dense Passage Retrieval}

\author[1,2]{Zhengyang Tang}
\author[2]{Benyou Wang}
\author[1]{Ting Yao}
\affil[1]{Tencent}
\affil[2]{The Chinese University of Hong Kong, Shenzhen}
\affil[ ]{ \texttt{\{zhytang,tessieyao\}@tencent.com, wangbenyou@cuhk.edu.cn} }

\begin{document}
\maketitle
\begin{abstract}
Deep prompt tuning (DPT) has gained great success in most natural language processing~(NLP) tasks. However, it is not well-investigated in dense retrieval where fine-tuning~(FT) still dominates. When deploying multiple retrieval tasks using the same backbone model~(e.g., RoBERTa), FT-based methods are unfriendly in terms of deployment cost:  each new retrieval model needs to repeatedly deploy the backbone model without reuse. To reduce the deployment cost in such a scenario, this work investigates applying DPT in dense retrieval. The challenge is that \textit{directly applying DPT in dense  retrieval largely underperforms FT methods}. To compensate for the performance drop,  we propose two model-agnostic and task-agnostic strategies for DPT-based retrievers, namely \textit{retrieval-oriented intermediate pretraining} and \textit{unified negative mining}, as a general approach that could be compatible with any pre-trained language model and retrieval task. The experimental results~\footnote{Our code is available at \url{https://github.com/tangzhy/DPTDR}} show that the proposed method (called DPTDR) outperforms previous state-of-the-art models on both MS-MARCO and Natural Questions. We also conduct ablation studies to examine the effectiveness of each strategy in DPTDR. We believe this work facilitates the industry, as it saves enormous efforts and costs of deployment and increases the utility of computing resources.

\end{abstract}

\section{Introduction}

Fine-tuning (FT) has been a de facto approach for effective dense passage retrieval~\citep{karpukhin2020dense,xiong2020approximate} based on pre-trained language models~(PLM). However, FT is unfriendly for industrial deployment in multi-task scenarios. Imaging for cloud service providers or infrastructure teams of search companies, each retrieval model (w.r.t., an individual task) necessarily re-deploys a backbone model since the weights of the backbone model in each task are fine-tuned and therefore slightly different. That dramatically increases hardware costs and inefficiency.

Recently, prompt tuning (PT)~\citep{liu2021pre} is a lightweight alternative to FT, which does not need storing a full copy of the backbone model for each task. One variant of PT, namely Deep Prompt Tuning (DPT; \citealp{li2021prefix,liu2021p}), exhibits comparable performances with FT in various NLP tasks. DPT enjoys parameter-efficient\citep{houlsby2019parameter} characteristics, of which the resulting prompts are light-weighted and can be easily passed to an online PLM service, thus overcoming the above challenge of FT. This paper asks: \textit{whether can we replace FT by DPT with comparable performance to SOTA FT methods in dense passage retrieval?} With comparable performance, DPT is much more friendly in deployment than FT.

DPT usually freezes weights in the backbone models and alternatively trains deep prompts inserted; the latter has much fewer parameters than the former. However, freezing most weights in DPT hinders its adaptability and therefore possibly harms performance.  Experimental results in Sec.~\ref{dptvsft} also demonstrate \textit{directly applying DPT in dense retrieval largely underperforms FT methods}.

To make DPT comparable to FT in dense retrieval, a natural solution is \textit{retrieval-oriented intermediate pretraining (RIP)}, which warms up the text representation via contrastive learning. Though it is not a novel idea\citep{lee2019latent,gao2021unsupervised,izacard2021towards}, there exist two different pretraining ways tailored for DPT-based retrievers. One is to pre-train deep prompts while freezing the PLM backbone and use the pre-trained prompts to initialize a DPT retriever. The other is to pre-train a PLM directly and initialize a DPT retriever using the pre-trained PLM; in contrast to prior works\citep{gao2021unsupervised}, we intend to allow any PLM easily pre-trained for DPT so that users may employ their own PLMs, and thus we deliberately remove the workload to modify any model structures. Surprisingly, empirical findings in Sec.~\ref{uscl} show that this choice yields better performance than carefully modified PLMs\citep{gao2021unsupervised}. Furthermore, we propose a \textit{unified negative mining (UNM)} to merge retrieved negatives from many retrievers including BM25 and dense retrievers, in order to provide diverse and hard negatives for DPT training. 

By incorporating RIP and UNM, we implement a \textbf{D}eep \textbf{P}rompt  \textbf{T}uning  method in \textbf{D}ense \textbf{R}etrieval tasks, called \emph{DPTDR}.
The experimental results show that DPTDR outperforms previous state-of-the-art models on both MS-MARCO and Natural Questions. We also conduct extensive experiments and find that: i) when combined with RIP and UNM, DPT is able to obtain comparable performance with FT in dense retrieval and exhibits insensitivity to prompt length, and ii) both RIP and UNM are effective in improving the performance. The contributions of this paper can be summarized as follows:

\begin{itemize}
  \item To our best knowledge, this is the first work to apply DPT in dense retrieval. We bring forward two essential strategies, namely retrieval-oriented intermediate pretraining and unified negative mining, allowing DPT to match FT's performance and be compatible with any PLM.
  \item Experiments show that DPTDR outperforms previous state-of-the-art models on MS-MARCO and Natural Questions and examine the effectiveness of the above strategies. 
  \item We believe this work facilitates the industry, as it saves enormous efforts and costs of deployment and increases the utility of computing resources.
\end{itemize}

\section{Related Work}

\subsection{Deep Prompt Tuning}

DPT originates from prompting and prompt tuning~\citep{liu2021pre}. Given some discrete or continuous prompts, PLMs like GPT-3\citep{brown2020language} can achieve impressive zero-shot and few-shot performances for knowledge-intensive tasks. However, studies find that prompt tuning fails to perform well for moderate-size models~\citep{liu2021p}. Thus, DPT\citep{li2021prefix,liu2021p} is proposed by inserting prompts at deep layers to steer PLMs towards desired directions more capably. It obtains comparable performance to FT across a range of NLP tasks. DPTDR is mainly related to DPT, focusing on dense passage retrieval instead of NLP. There also exist works of pretraining prompts for prompt tuning\citep{gu2021ppt}, which shows effectiveness in few-shot learning using billion-size models, as we will explore as well in the context of DPT.

\subsection{Dense Retrieval}
\textbf{Pretraining} We have witnessed a series of unsupervised pretraining works proposed for dense retrieval, such as ICT, BFS, WLP, and independent cropping~\citep{lee2019latent, chang2020pre, izacard2021towards}. Following works also try to pre-train retriever and reader jointly for question answering~\citep{guu2020realm}. coCondenser~\citep{gao2021unsupervised} follows a contrastive learning framework using Condenser structure~\citep{gao2021condenser} by adding an explicit decoder to learn representations better. There are also semi-supervised and weakly-supervised works. DPR-PAQ~\citep{ouguz2021domain} pre-trains a PLM using 65-million-size synthetic QA pairs on the target corpus. GTR~\citep{ni2021large} pre-trains T5~\citep{raffel2019exploring} on 2-billion-size community QA pairs from T5-base to T5-xxlarge. We follow unsupervised contrastive learning as our pretraining strategy for DPTDR. However, we aim to ensure compatibility with any PLM, thus resulting in different sample building processes and model structure choices.

\textbf{Negative mining} DPR~\citep{karpukhin2020dense} proposes to train retrievers using BM25 negatives. ANCE~\citep{xiong2020approximate} extends that by mining negatives periodically from previously-trained dense retrievers. RocketQA and RocketQAv2~\citep{qu2021rocketqa,ren2021rocketqav2} introduce the idea of denoised negative sampling by selecting negatives with high confidence scored by a re-ranker. DPTDR unifies the above into a general negative mining strategy.

\section{Methodology} \label{section:3}

\begin{figure*}[h]
\centering
\includegraphics[width=1\textwidth]{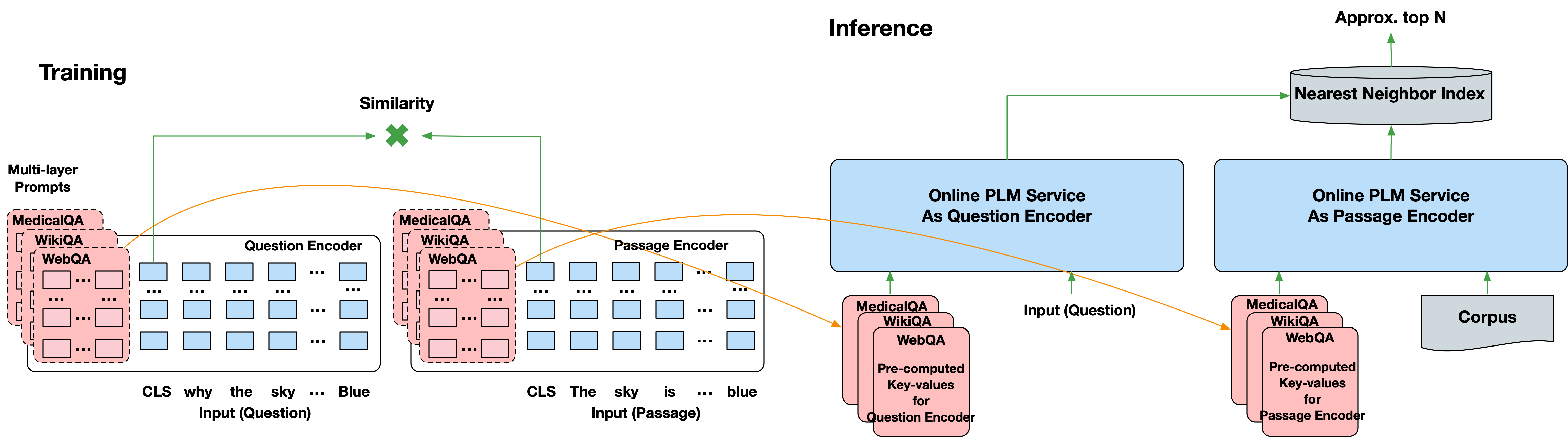}
  \caption{The framework of DPTDR. We first perform RIP which results in a PLM (the blue blocks) that can be used as the backbone for DPT training and deployed once as online PLM services. Then we train deep prompts (i.e., DPT) for different retrieval tasks such as WebQA, WikiQA, and MedicalQA (the pink blocks), during which we may employ UNM to improve performances. For inference, we can send tokenized input, together with trained prompts of their corresponding task, to online PLM services to get dense vectors.}
  \label{fig:dptdr}
\end{figure*}

In this section, we first formalize the application of DPT in dense retrieval. We then describe the two strategies of RIP and UNM for DPT-based retrievers.

\subsection{DPT in Dense Retrieval}
Let $C$ be a corpus consisting $N$ passages, denoted by $p_1, p_2, ..., p_N$. Given a question $q$, the task of dense retrieval is to find a passage $p_i$ that is considered relevant to the question. 

\textbf{The dual-encoder} Normally, a dual-encoder is applied. First its passage encoder $E_p(\cdot)$ embeds a passage $p$ to a $d$-dimensional dense vector. Then a vector search index~\citep{johnson2019billion} of passages is built for retrieval. At inference time, the question encoder $E_q(\cdot)$ embeds the question $q$ to a $d$-dimensional dense vector, and $k$ passages closet to the question based on the vector similarity will be retrieved. In practice, the similarity score is computed as the inner product:
\begin{equation}
s(q, p) = E_q(q) \cdot E_p(p).
\end{equation}

For PLM-based dual-encoder, we usually take the representation at the first token (e.g., [CLS] symbol in BERT~\citep{devlin2018bert}) as the output dense vector.

\textbf{Deep prompt tuning} We then apply DPT in the PLM-based dual-encoder, as illustrated in the left part of Figure~\ref{fig:dptdr}. To prepend multi-layer prompts for the dual-encoder, we initialize a trainable prefix matrix $M$ of dimension $l \times d$ for each layer of the PLM, where $l$ is the length of the prompt and $d$ is the hidden size of the PLM. Since the prompt resides at the deep layers of PLM, it has a full capacity to steer the PLM towards the desired direction and output meaningful dense vector for questions and passages. Note that a verbalizer~\citep{schick2020s} plays a vital role in mapping words to labels in canonical prompt tuning. However, we remove it in dense retrieval since the output dense vector is what we need. Let $E_p'$ as the prompted passage encoder and $E_q'$ as the prompted question encoder, and the similarity score is computed:
\begin{equation}\label{eq2}
s'(q, p) = E_q'(q) \cdot E_p'(p).
\end{equation}

\textbf{Training}
The objective of the training is to learn dense vectors so that the similarity between relevant pairs of questions and passages ranks higher than irrelevant ones. Given a pair of question $q$ and positive passage $p_i$, along with $n$ negative passages, we optimize the loss function as the negative log-likelihood of the positive passage: 

\begin{multline}
L(q_i, p_i^+, \{p_{i,j}^-\}_{j=1}^n) = \\
- \log \frac{e^{s'(q_i, p_i^+)}}{e^{s'(q_i, p_i^+)} + \sum_{j=1}^{n} e^{s'(q_i, p_{i,j}^-)}}.
\end{multline}

Generating negative passages is critical for the performance, and we will explain it in Sec. \ref{unm}. During training, we freeze parameters of the backbone PLM and only update the deep prompts, where approximately 0.1\%-0.4\% parameters of a PLM get trained. 

\textbf{Inference} As illustrated in the right part of Figure~\ref{fig:dptdr}, since the backbone PLM is frozen, it is possible to deploy it ahead as online PLM services and then pass the trained prompts as pre-computed key values together with tokenized inputs to get dense vectors. It is at the core of how we save efforts and costs of deployment and increase the utility of computing resources. In practice, the cloud service providers or infrastructure teams of search companies are able to focus on the PLM as a central service, while users can quickly train deep prompts for different retrieval tasks and obtain efficient and compelling retrieval performances without any deployment. 

Although DPT brings in many advantages, it is worth noting that it does not accelerate the inference speed because the forward computation is not reduced but increased slightly. 

\subsection{Retrieval-oriented Intermediate Pretraining (RIP) for DPT}\label{sec:pretrain}
The goal of RIP is to either pre-train deep prompts or PLMs using contrastive learning. We first describe the task as follows. Let $C$ denote a corpus consisting $N$ passages. For a passage $p_i$, we split it into $l$ sentences, denoted by $s_i^1, ..., s_i^l$. Given a sentence $s_i^j$, the task of pretraining is to distinguish its context sentence $s_i^{j'}$ from sentences of other passages $s_k^{l}$, where $k \neq i$. Formally, we randomly select a pair of sentences from each passage as context sentences to form a batch of training data $B = \{s_i^1, s_i^2\}_{i=1}^{m}$, where $m$ is the batch size. Then we define the contrastive loss for $s_i^j$ over the batch as:

\begin{equation}
L_{c}(s_i^j) = 
- \log \frac{ e^{s(s_i^1, s_i^2)} }{ \sum_{k=1}^{m} \sum_{l=1}^{2} {_{{ij} \neq {kl}}} e^{s(s_i^j, s_k^l)} }. %
\end{equation}

In contrast to prior works\citep{gao2021unsupervised, izacard2021towards}, we directly sample sentences as opposed to text spans. Since sampling text spans is a non-trivial technique where factors such as the probability of short sentences and how to keep the spans linguistically meaningful can have a complicated effect on the pretraining, we remove this complexity in our approach. We also conduct an experiment observing sentences work even better than text spans in MS-MARCO corpus (Sec. \ref{sec:spanvssent}).

Under the contrastive learning task, there exist two pretraining ways tailored for DPT, depending on the pre-trained objects (i.e., the deep prompts or the PLM backbone).

\textbf{Pre-train deep prompts} One is to pre-train deep prompts with a vanilla PLM. Later we initialize a DPT-based retriever using the pre-trained deep prompts and the vanilla PLM. However, experiments in Sec. \ref{sec:ppt} show that it suffers from catastrophic forgetting and exhibits no superior performance to randomly initialized prompts.

\textbf{Pre-train the PLM} The other is to pre-train a PLM, and then we initialize a DPT-based retriever using randomly-initialized deep prompts and the pre-trained PLM. Notice that we intend to allow any PLM to be easily pre-trained for DPT so that users may employ their own PLMs. Thus we contrast prior works such as coCondenser\citep{gao2021unsupervised}, a state-of-the-art model structure in contrastive pretraining, by removing the workload to modify any model structures. Surprisingly, it yields better performance than coCondenser in Table~\ref{tab:plm}. Therefore, we refer RIP strategy as pretraining of PLMs for the rest.

For any PLM, We also intend to remain its original self-supervised tasks, such as masked language modeling(MLM; \citealp{devlin2018bert,sun2019ernie}), denoted as $L_{s}$. Therefore, the final loss of pretraining over the batch is:
\begin{equation}
L = \frac{1}{2m} \sum_{i=1}^{m} \sum_{j=1}^{2} L_{s}(s_i^j) + L_{c}(s_i^j).
\end{equation}

After pretraining, the resulting model can be deployed once as online services and taken as the backbone model for DPT training.

\subsection{Unified Negative Mining (UNM)}\label{unm}

We also develop unified negative mining for DPT, as interpreted as "Multiple Retrievers \& Hybrid Sampling." "Multiple Retrievers" is to incorporate negatives from as many retrievers as we can. We use a BM25 retriever as the initial retriever and train a DPT-based retriever using BM25 negatives. Later we treated retrieved negatives from the BM25 retriever and the first DPT-based retriever as un-denoised hard negatives. Users are allowed to introduce any other retrievers if possible. "Hybrid Sampling" is to select denoised hard negatives from un-denoised hard negatives retrieved by the above multiple retrievers. We borrow an existing re-ranker released by RocketQA~\citep{qu2021rocketqa} and select those negatives with high confidence. For training the final DPT-based retriever, we mix the denoised hard negatives, un-denoised hard negatives, and easy negatives from in-batch or cross-batch training. 

We believe unified negative mining is critical for the performance of DPT-based retrievers, as it provides negatives of high quality and diversity.

\section{Experiments}



\begin{table*}[h]
\centering
  \caption{The statistics of MS-MARCO and Natural Questions.}
  \label{tab:data}
  \scalebox{0.9}{
  \begin{tabular}{lcccc}
    \toprule
    \textbf{Dataset} & \textbf{\#q in train} & \textbf{\#q in dev} & \textbf{\#q in test} & \textbf{\#passages} \\
    \midrule
    MS-MARCO & 502,939 & 6,980 & 6,837 & 8,841,823 \\
    Natural Questions & 58,812 & 6,515 & 3,610 & 21,015,324 \\
    \bottomrule
  \end{tabular}
  }
\end{table*}

\subsection{Experimental Setting}
\paragraph{Datasets and metrics}
We experiment with two popular dense retrieval datasets, including MS-MARCO~\citep{bajaj2016ms} and Natural Questions(NQ; \citealp{karpukhin2020dense}). The statistics of the datasets are listed in Table~\ref{tab:data}. MS-MARCO is constructed from Bing’s search query logs and web documents retrieved by Bing. Natural Question contains questions from Google Search. For evaluation, we report official metrics MRR@10, RECALL@1000 for MS-MARCO, and RECALL at 5, 20, and 100 for NQ. All models are trained on a single server with 8 NVIDIA Tesla A100 GPUs.

\paragraph{Settings in DPT}
We use RoBERTa-large-size models as the backbone for DPT training. Hyper-parameters are explored as below.
\begin{itemize}
    \item \textbf{Learning rate} We search for {1e-2, 5e-3, 7e-3, 5e-4, 5e-5, 5e-6} with prompts' length of 32, where 7e-3 performs relatively better than others and is set for the main experiment.
    \item \textbf{Training epochs} For training epochs, we search for {3, 6, 10} with a learning rate 7e-3 on MS-MARCO, where 10 performs best and is set for the main experiment. We also set training epochs as 60 for NQ for acceptable time cost.
    \item \textbf{Prompt length} We search for {8, 16, 32, 64, 128}, as is discussed in Sec.~\ref{promptlength}. We use 128 for the main experiment. 
    \item \textbf{Reparametrization} We also conduct experiments for prompts with or without MLP reparametrization, as is discussed in Sec.~\ref{reparam}. We use non-reparametrization for the main experiment.
\end{itemize}
We follow  coCondenser~\citep{gao2021unsupervised} for other hyper-parameters (e.g., parameter sharing, batch size, warm-up ratio, and mixed-precision training).

\paragraph{Settings in RIP}\label{exp:pretrain}
We choose to pre-train vanilla RoBERTa-large for RIP, whose model size appears more common for DPT~\citep{li2021prefix,liu2021p} and is consistent with the above DPT training. We remain RoBERTa's original self-supervised task (MLM;~\citealp{liu2019roberta}). To compare our approach with coCondenser~\citep{gao2021unsupervised}, we also pre-train a coCondesner RoBERTa-large. Since coCondenser modifies the PLM by adding a carefully designed Condenser structure, we follow their structural setting using an equal split, 12 early layers, and 12 late layers. We split the passages into sentences on both MS-MARCO and NQ Wikipedia as the training corpus. The models are trained using AdamW optimizer with a learning rate 1e-4, weight decay of 0.01, linear learning rate decay, and a batch size of 2K. We train 8 epochs for MS-MARCO and 4 epochs for NQ Wikipedia. 

\paragraph{Settings in UNM} For un-denoised hard negatives, we randomly select 30 out of the top 200 retrieved negatives from multiple retrievers. For denoised hard negatives, we select negatives with a score less than 0.1 output by an existing re-ranker~\citep{qu2021rocketqa}.

\begin{table*}[h]
\centering
  \caption{Passage retrieval results on MS-MARCO Dev and Natural Questions Test. We copy the results from the original papers. The best and second-best results are in bold and underlined fonts respectively.}
  \label{tab:results}
  \scalebox{0.9}{
  \begin{tabular}{llccccc}
    \toprule
    \multirow{2}{1em}{\textbf{Methods}} & \multirow{2}{1em}{\textbf{PLM}} & \multicolumn{2}{c}{\textbf{MS-MARCO Dev}} & \multicolumn{3}{c}{\textbf{Natural Questions Test}} \\
     &  & MRR@10 & R@1000 & R@5 & R@20 & R@100 \\
    \midrule
    BM25 & - & 18.7 & 85.7 & - & 59.1 & 73.7 \\
    DeepCT\citep{dai2019deeper} & - & 24.3 & 91.0 & - & - & -  \\
    docT5query\citep{nogueira2019doc2query} & - & 27.7 & 94.7 & - & - & - \\
    GAR\citep{mao2020generation} & - & - & - & - & 74.4 & 85.3 \\
    \hline
    DPR\citep{karpukhin2020dense} & BERT-base & - & - & - & 78.4 & 85.4 \\
    ANCE\citep{xiong2020approximate} & RoBERTa-base & 33.0 & 95.9 & - & 81.9 & 87.5 \\
    ME-BERT\citep{luan2020sparse} & BERT-large & 34.3 & - & - & - & - \\
    RocketQA\citep{qu2021rocketqa} & ERNIE-base & 37.0 & 97.9 & 74.0 & 82.7 & 88.5 \\
    RocketQAv2\citep{ren2021rocketqav2} & ERNIE-base & \underline{38.8} & 98.1 & 75.1 & 83.7 & 89.0 \\
    \hline
    coCondenser\citep{gao2021unsupervised} & Condenser & 38.2 & 98.4 & 75.8 & 84.3 & 89.0 \\
    \hline
    DPR-PAQ\citep{ouguz2021domain} & RoBERTa-large & 34.0 & - & \underline{76.9} & \underline{84.7} & \underline{89.2} \\
    \hline
    \multirow{4}{12em}{GTR\citep{ni2021large}} & T5-base & 36.6 & 98.3 & - & - & - \\
     & T5-large & 37.9 & \textbf{99.1} & - & - & - \\
     & T5-xlarge & 38.5 & 98.9 & - & - & - \\
     & T5-xxlarge & \underline{38.8} & \underline{99.0} & - & - & - \\
    \hline
    DPTDR & RoBERTa-large & \textbf{39.1} & 98.9 & \textbf{77.5} & \textbf{85.1} & \textbf{89.4} \\
    \bottomrule
  \end{tabular}
  }
\end{table*}

\paragraph{Baselines}
We use the following baselines.
\textbf{coCondenser} ~\citep{gao2021unsupervised}  designs a complicated pretraining model structure on top of a vanilla PLM.
\textbf{DPR-PAQ }~\citep{ouguz2021domain} pre-trains a RoBERTa-large using 65-million-size synthetic QA pairs. Since the data is created by a model trained on NQ~\citep{kwiatkowski2019natural} and Trivia QA~\citep{joshi2017triviaqa}, it can be considered a semi-supervised method. It is also comparable to us as both of us use RoBERTa-large.
\textbf{GTR} ~\citep{ni2021large} pre-trains T5 encoder ~\citep{raffel2019exploring} using 2-billion size community QA pairs. It also provides results across all model size ranges from T5-base to T5-xxlarge. The massive training corpus and model size establish a SOTA performance.

We also include some standard baselines including sparse retrieval systems (BM25, DeepCT~\citep{dai2019deeper}, DocT5Query~\citep{nogueira2019doc2query}, and GAR~\citep{mao2020generation}) and dense retrieval systems ( DPR~\citep{karpukhin2020dense}, ANCE~\citep{xiong2020approximate}, ME-BERT~\citep{luan2020sparse}, and RocketQA~\citep{qu2021rocketqa}). We also include RocketQAv2~\citep{ren2021rocketqav2} as it jointly trains the retriever and reranker using hybrid sampled negatives.


\subsection{Experimental Results}
\subsubsection{Comparison with Existing Methods}
Table~\ref{tab:results} shows the dev set performance for MS-MARCO and test set performance for NQ. We can generally see that \textbf{DPTDR outperforms all the baselines in terms of MRR@10 on MS-MARCO and R@5 on NQ} and set a new SOTA in the two datasets.

We first compare DPTDR with DPR-PAQ. DPR-PAQ achieves competitive performance on NQ. It should be expected since it involves large semi-supervised pretraining on the NQ dataset. Nonetheless, DPTDR still outperforms DPR-PAQ by 0.6 points in R@5 although we use an unsupervised pretraining model. When we study the performance on MS-MARCO, DPR-PAQ fails to perform as consistently well as on NQ, which could result from domain mismatch of pretraining, and DPTDR outperforms it by a significant margin of 5.1 points in MRR@10.

Secondly, we compare DPTDR with GTR. GTR pre-trains T5 using 2-billion-size community QA pairs as a weakly-supervised pretraining. For such a scale of training corpus, we would expect that larger models would consume the corpus more thoroughly and perform better on downstream tasks. As a result, GTR consistently boosts the performance on MS-MARCO with the model size increasing. However, DPTDR still outperforms GTR T5-xxlarge, a 10-billion-size model, and outperforms GTR T5-large by a noticeable margin of 1.2 points in MRR@10. It shows that model size is a positive contributor but not an absolute dominator for dense retrieval. Appropriate pretraining and negative mining can help improve performances using much more affordable computing resources. At the same time, note that DPT shall play a critical role in achieving comparable performance to FT with the help of RIP and UNM. We will validate this in Sec.~\ref{dptvsft}.

Finally, we would like to compare DPTDR with coCondenser. Since coCondenser employs a pre-trained Condenser model\citep{gao2021condenser}, we will conduct a more fair comparison in Sec.~\ref{uscl}.

\subsubsection{Comparing FT with and without RIP and UNM Strategies} \label{dptvsft}
To answer the raised question: \textit{whether can we replace FT by DPT with comparable performance to SOTA FT methods in dense passage retrieval?} We conduct FT by following hyper-parameters of coCondenser~\citep{gao2021unsupervised}.

\begin{table*}[h]
\centering
  \caption{The comparison between FT and DPT with and without RIP and UNM strategies on MS-MARCO Dev and Natural Questions Test. \textbf{DPT with  RIP\&UNM} is the proposed method, a.k.a, `DPTDR'.}
  \label{tab:ftvsdpt}
  \scalebox{0.9}{
  \begin{tabular}{ll|ll|lll}
  \toprule
    \multirow{2}{1em}{\textbf{}} & \multirow{2}{1em}{\textbf{}} & \multicolumn{2}{c}{\textbf{MS-MARCO Dev}} & 
    \multicolumn{3}{c}{\textbf{Natural Questions Test}}\\
    &   &   MRR@10 & R@1000 & R@5 & R@20 & R@100 \\
    \midrule
     \multirow{2}{7em}{{ w/o RIP\&UNM}} & FT & 34.9 & 97.2 & 68.8 & 80.0 & 86.4 \\
     & DPT & 32.7 ( \textbf{2.2} $ \downarrow$) & 96.3 (\textbf{0.9} $ \downarrow$) & 66.5 ( \textbf{2.3} $ \downarrow$) & 78.5 ( \textbf{1.5} $ \downarrow$) & 85.5 ( \textbf{0.9} $ \downarrow$)\\ 
     \hline
      \multirow{2}{7em}{{ w/ RIP\&UNM}} & FT  & 39.4 & 99.0 & 77.0 & 85.4 & 89.2 \\
      &DPT   & 39.1 ( \textbf{0.3} $ \downarrow$) & 98.9 (\textbf{ 0.1} $ \downarrow$) & 77.5 ( \textbf{0.5} $ \uparrow$) & 85.1 ( \textbf{0.3} $ \downarrow$) & 89.4 ( \textbf{0.2} $ \uparrow$)\\
    \bottomrule
  \end{tabular}
  }
\end{table*}

\paragraph{Comparison w/o RIP\&UNM}
As a starter, we examine the effectiveness of directly replacing FT with DPT, which means we conduct training without RIP and UNM strategies. Thus we use the vanilla RoBERTa-large as the backbone model and BM25 negatives. As is shown in Table~\ref{tab:ftvsdpt}. We notice that DPT largely underperforms FT in this setting with a noticeable margin of 2.2 points in MRR@10 on MS-MARCO and 2.3 points in R@5 on NQ. It indicates that freezing most weights in DPT actually hinders its adaptability and therefore harms performance.

\paragraph{Comparison w/ RIP\&UNM}
Next, we examine the performance of FT and DPT with RIP and UNM strategies. We use the RIP RoBERTa-large as the backbone model and UNM negatives. Table~\ref{tab:ftvsdpt} shows that \textit{i}) RIP and UNM improve the performances of both FT and DPT and \textit{ii}) most importantly, DPT is comparable to FT under this setting, where the gap shrinks to only 0.3 points in MRR@10 on MS-MARCO, and DPT even slightly outperforms FT by 0.5 points in R@5 on NQ. As a result, we can see that when combined with RIP and UNM, DPT can obtain comparable performance with FT in dense retrieval.


\subsection{Analysis on DPT}
 \label{promptlength}

\paragraph{Sensitivity on prompt length}
We also seek to understand how prompt length affects the performance of DPT-based retrievers. From Table~\ref{tab:promptlength}, we observe that the performance of prompt length of 8 already achieves a strong MRR@10 at $38.6$ on MS-MARCO. When we increase the length to 128, it makes the most robust performance of MRR@10 at $39.1$. The longer prompt means more trainable parameters, which obtains more power to steer PLMs. However, we also want to point out that the DPT retriever exhibits insensitivity to prompt length since the performances are competitive overall across various lengths. Therefore, we choose 32 as the default prompt length along with other hyper-parameters in the main experiment for the rest of the ablation studies on MS-MARCO to accelerate the training.

\begin{table}[h]
\centering
  \caption{Sensitivity of prompt length on MS-MARCO Dev.}
  \label{tab:promptlength}
  \scalebox{0.9}{
  \begin{tabular}{lcc}
    \toprule
    \textbf{Prompt Length} & MRR@10 & R@1000 \\
    \midrule
    8 & 38.6 & 98.9 \\
    16 & 38.6 & 99.0 \\
    32 & 38.7 & 98.9 \\
    64 & 38.5 & 98.9 \\
    128 & 39.1 & 98.9 \\
    \bottomrule
  \end{tabular}
  }
\end{table}

\paragraph{Impact of reparameterization} \label{reparam}
Reparametrization plays an important role in DPT. \citealp{li2021prefix} point out that MLP reparametrization results in more stable and compelling performances, while \citealp{liu2021p} find it still depends on different tasks. In dense retrieval, we aim to determine whether it has a positive effect. Table~\ref{tab:reparam} presents the results on MS-MARCO. We observe that MLP reparametrization results in a performance drop in MRR@10 on MS-MARCO. Since MLP breaks the independence of inter-layer prompts, we conjecture this brings optimization difficulty for dense retrieval.

\begin{table}[h]
\centering
  \caption{Ablations of reparamerization on MS-MARCO Dev.}
  \label{tab:reparam}
  \scalebox{0.9}{
  \begin{tabular}{lcc}
    \toprule
    \textbf{Reparamerization} & MRR@10 & R@1000 \\
    \hline
    embedding & 38.7 & 98.9 \\
    mlp & 38.0 & 99.0 \\
    \bottomrule
  \end{tabular}
  }
\end{table}

\subsection{Analysis  on RIP}

\paragraph{Whether to pre-train deep prompts or not?}
\label{sec:ppt}
We try to examine whether pre-trained deep prompts could improve the performance of DPT-based retrievers. We use BERT-base as our backbone model and pre-train deep prompts of length 32 without reparameterization. The pretraining tasks and corpus are exactly the same as Sec.~\ref{sec:pretrain}. We initialize DPT-based retrievers using pre-trained and randomly-initialized prompts. As is shown in Table \ref{tab:ppt}, the pre-trained prompts do not boost the performance over randomly initialized prompts on MS-MARCO. It reveals that the deep prompts may easily suffer from catastrophic forgetting.

\begin{table}[h]
\centering
  \caption{Ablations of prompt initialization on MS-MARCO Dev.}
  \label{tab:ppt}
  \scalebox{0.9}{
  \begin{tabular}{lcc}
    \toprule
    \textbf{Prompt Initialization} & MRR@10 & R@1000 \\
    \hline
    Random & 32.4 & 95.5 \\
    Pre-trained & 32.4 & 95.5 \\
    \bottomrule
  \end{tabular}
  }
\end{table}

\paragraph{RIP on text spans or sentences}
\label{sec:spanvssent}
We also explore pretraining using randomly-sampled sentences versus randomly-sampled text spans. Since coCondenser\citep{gao2021unsupervised} releases their pre-trained model using randomly-sampled text spans, we directly use their model to examine the zero-shot performance. For sampling sentences, we use the same PLM and hyper-parameters based on coCondenser code\footnote{\url{https://github.com/luyug/Condenser}} except changing the training corpus consisting of randomly-sampled sentences. Table~\ref{tab:spanvssent} presents the zero-shot performance on MS-MARCO. The pretraining using sentences works better than the one using text spans. This is might be owing to that text-spans based RIP does not consider the (starting and ending) borders  of natural sentences and therefore break their completeness in semantics.

\begin{table}[h]
\centering
\addtolength\tabcolsep{-4pt}
  \caption{Zero-shot performance of coCondenser with different sampling granularity (i.e., sentences or spans) on MS-MARCO Dev.}
  \label{tab:spanvssent}
  \scalebox{0.9}{
  \begin{tabular}{lcc}
    \toprule
    \textbf{Unit} & MRR@10 & R@1000 \\
    \midrule
    Spans~\citet{gao2021unsupervised} & 11.1 & 78.2 \\
    Sentences & 15.4 & 87.2 \\
    \bottomrule
  \end{tabular}
  }
\end{table}

\paragraph{RIP's effectiveness and comparison with coCondenser} \label{uscl}

\begin{table*}[h]
\centering
  \caption{Ablations of different PLMs for DPT on MS-MARCO Dev.}
  \label{tab:plm}
  \scalebox{0.9}{
  \begin{tabular}{l|cccc|cc}
    \toprule
    \multirow{2}{4em}{\textbf{Backbone PLM}} & \multicolumn{4}{c}{\textbf{Zero-shot}} & 
    \multicolumn{2}{c}{\textbf{Full-shot}} \\
    & $l_{align}$ & $l_{uniform}$ & MRR@10 & R@1000 & MRR@10 & R@1000 \\
    \midrule
    vanilla RoBERTa-large & 161.4 & -13.8 & 0.0 & 0.1 & 35.5 & 97.5 \\
    coCondenser RoBERTa-large & 4.9 & -12.9 & 6.4 & 63.3 & 37.3 & 98.0 \\
    RIP RoBERTa-large & 21.9 & -26.4 & 14.3 & 87.2 & 38.7 & 98.9 \\
   \bottomrule
  \end{tabular}
  }
\end{table*}

We also try to examine the effectiveness of RIP strategy and compare it with coCondenser~\citep{gao2021unsupervised}. Concretely, we take vanilla RoBERTa-large, coCondenser RoBERTa-large, and RIP RoBERTa-large as the backbone model for DPT training under the same setting. Table~\ref{tab:plm} presents their results in both zero-shot and full-shot settings on MS-MARCO. For vanilla RoBERTa-large, it performs extremely poorly in zero-shot experiments, and with no surprise, it performs worst in full-shot experiments among the three PLMs. For coCondenser RoBERTa-large, it achieves a noticeable improvement over vanilla RoBERTa-large, where MRR@10 of zero-shot performance becomes meaningful at 6.3, and MRR@10 of full-shot performance increases to 37.3. For RIP RoBERTa-large, we see it achieves the best performance in both zero-shot and full-shot experiments. We also borrow the analysis tool from~\citet{wang2020understanding}, which takes $l_{align}$ between semantically-related positive pairs and $l_{uniform}$ of representation space to measure the quality of PLM representations. For both the metrics, lower numbers are better. RIP is much better than the vanilla model in both alignment and uniformity, while coCondenser works well in alignment but worse in uniformity. 

Thus a question is raised: \textit{does PLM need additional structures for contrastive pretraining?} Both zero-shot and full-shot experiments demonstrate that RIP works even better than a carefully modified model structure. Therefore, we conjecture that PLM's multi-layer transformers could be already expressive enough for dense retrieval under an appropriate contrastive learning task. However, additional model structures may bring  optimization difficulty, especially when the number of added parameters is large.

\subsection{Analysis  on UNM}

\paragraph{Ablation on UNM}
We try to understand how UNM affects performances. Table~\ref{tab:unified} presents the results on MS-MARCO. DPT using BM25 negatives achieves a baseline of MRR@10 at 36.8. When combining un-denoised hard negatives from multiple retrievers, we see that the performance achieves a noticeable improvement in MRR@10 by 1.5 points. When combining denoised hard negatives selected by a re-ranker, the performance further gets boosted of which MRR@10 increases by 0.4 points. The results demonstrate that both multiple retrievers and hybrid sampling positively contribute to dense retrieval.

\begin{table}[h]
\centering
  \caption{Ablations of UNM on MS-MARCO Dev.}
  \label{tab:unified}
  \scalebox{0.9}{
  \begin{tabular}{lcc}
   \toprule
    \textbf{Neg Pool} & MRR@10 & R@1000 \\
    \midrule
    BM25 Neg & 36.8 & 98.6 \\
    \quad + un-denoised   Neg & 38.3 & 98.9 \\
     \quad + denoised   Neg & 38.7 & 98.9 \\
   \bottomrule
  \end{tabular}
  }
\end{table}

\section{Conclusion}
In this paper, we investigate applying DPT in dense passage retrieval. To mitigate the performance drop of a vanilla DPT, We also propose two strategies, namely RIP and UNM, to enhance DPT and match the performance of FT. Experiments show that DPTDR outperforms previous state-of-the-art models on both MS-MARCO and Natural Questions and demonstrated the effectiveness of the above strategies. We believe this work facilitates the industry, as it saves enormous efforts and costs of deployment and increases the utility of computing resources. In future work, we will explore scaling up the model size to further improve DPTDR.

\section*{Acknowledgment}

Benyou Wang is funded by the CUHKSZ startup funding   No. UDF01002678.
\bibliography{anthology,custom}

\begin{thebibliography}{32}
\expandafter\ifx\csname natexlab\endcsname\relax\def\natexlab#1{#1}\fi

\bibitem[{Bajaj et~al.(2016)Bajaj, Campos, Craswell, Deng, Gao, Liu, Majumder,
  McNamara, Mitra, Nguyen et~al.}]{bajaj2016ms}
Payal Bajaj, Daniel Campos, Nick Craswell, Li~Deng, Jianfeng Gao, Xiaodong Liu,
  Rangan Majumder, Andrew McNamara, Bhaskar Mitra, Tri Nguyen, et~al. 2016.
\newblock Ms marco: A human generated machine reading comprehension dataset.
\newblock \emph{arXiv preprint arXiv:1611.09268}.

\bibitem[{Brown et~al.(2020)Brown, Mann, Ryder, Subbiah, Kaplan, Dhariwal,
  Neelakantan, Shyam, Sastry, Askell et~al.}]{brown2020language}
Tom~B Brown, Benjamin Mann, Nick Ryder, Melanie Subbiah, Jared Kaplan, Prafulla
  Dhariwal, Arvind Neelakantan, Pranav Shyam, Girish Sastry, Amanda Askell,
  et~al. 2020.
\newblock Language models are few-shot learners.
\newblock \emph{arXiv preprint arXiv:2005.14165}.

\bibitem[{Chang et~al.(2020)Chang, Yu, Chang, Yang, and Kumar}]{chang2020pre}
Wei-Cheng Chang, Felix~X Yu, Yin-Wen Chang, Yiming Yang, and Sanjiv Kumar.
  2020.
\newblock Pre-training tasks for embedding-based large-scale retrieval.
\newblock \emph{arXiv preprint arXiv:2002.03932}.

\bibitem[{Dai and Callan(2019)}]{dai2019deeper}
Zhuyun Dai and Jamie Callan. 2019.
\newblock Deeper text understanding for ir with contextual neural language
  modeling.
\newblock In \emph{Proceedings of the 42nd International ACM SIGIR Conference
  on Research and Development in Information Retrieval}, pages 985--988.

\bibitem[{Devlin et~al.(2018)Devlin, Chang, Lee, and
  Toutanova}]{devlin2018bert}
Jacob Devlin, Ming-Wei Chang, Kenton Lee, and Kristina Toutanova. 2018.
\newblock Bert: Pre-training of deep bidirectional transformers for language
  understanding.
\newblock \emph{arXiv preprint arXiv:1810.04805}.

\bibitem[{Gao and Callan(2021{\natexlab{a}})}]{gao2021condenser}
Luyu Gao and Jamie Callan. 2021{\natexlab{a}}.
\newblock Condenser: a pre-training architecture for dense retrieval.
\newblock In \emph{Proceedings of the 2021 Conference on Empirical Methods in
  Natural Language Processing}, pages 981--993.

\bibitem[{Gao and Callan(2021{\natexlab{b}})}]{gao2021unsupervised}
Luyu Gao and Jamie Callan. 2021{\natexlab{b}}.
\newblock Unsupervised corpus aware language model pre-training for dense
  passage retrieval.
\newblock \emph{arXiv preprint arXiv:2108.05540}.

\bibitem[{Gu et~al.(2021)Gu, Han, Liu, and Huang}]{gu2021ppt}
Yuxian Gu, Xu~Han, Zhiyuan Liu, and Minlie Huang. 2021.
\newblock Ppt: Pre-trained prompt tuning for few-shot learning.
\newblock \emph{arXiv preprint arXiv:2109.04332}.

\bibitem[{Guu et~al.(2020)Guu, Lee, Tung, Pasupat, and Chang}]{guu2020realm}
Kelvin Guu, Kenton Lee, Zora Tung, Panupong Pasupat, and Ming-Wei Chang. 2020.
\newblock Realm: Retrieval-augmented language model pre-training.
\newblock \emph{arXiv preprint arXiv:2002.08909}.

\bibitem[{Houlsby et~al.(2019)Houlsby, Giurgiu, Jastrzebski, Morrone,
  De~Laroussilhe, Gesmundo, Attariyan, and Gelly}]{houlsby2019parameter}
Neil Houlsby, Andrei Giurgiu, Stanislaw Jastrzebski, Bruna Morrone, Quentin
  De~Laroussilhe, Andrea Gesmundo, Mona Attariyan, and Sylvain Gelly. 2019.
\newblock Parameter-efficient transfer learning for nlp.
\newblock In \emph{International Conference on Machine Learning}, pages
  2790--2799. PMLR.

\bibitem[{Izacard et~al.(2021)Izacard, Caron, Hosseini, Riedel, Bojanowski,
  Joulin, and Grave}]{izacard2021towards}
Gautier Izacard, Mathilde Caron, Lucas Hosseini, Sebastian Riedel, Piotr
  Bojanowski, Armand Joulin, and Edouard Grave. 2021.
\newblock Towards unsupervised dense information retrieval with contrastive
  learning.
\newblock \emph{arXiv preprint arXiv:2112.09118}.

\bibitem[{Johnson et~al.(2019)Johnson, Douze, and
  J{\'e}gou}]{johnson2019billion}
Jeff Johnson, Matthijs Douze, and Herv{\'e} J{\'e}gou. 2019.
\newblock Billion-scale similarity search with gpus.
\newblock \emph{IEEE Transactions on Big Data}.

\bibitem[{Joshi et~al.(2017)Joshi, Choi, Weld, and
  Zettlemoyer}]{joshi2017triviaqa}
Mandar Joshi, Eunsol Choi, Daniel~S Weld, and Luke Zettlemoyer. 2017.
\newblock Triviaqa: A large scale distantly supervised challenge dataset for
  reading comprehension.
\newblock \emph{arXiv preprint arXiv:1705.03551}.

\bibitem[{Karpukhin et~al.(2020)Karpukhin, O{\u{g}}uz, Min, Lewis, Wu, Edunov,
  Chen, and Yih}]{karpukhin2020dense}
Vladimir Karpukhin, Barlas O{\u{g}}uz, Sewon Min, Patrick Lewis, Ledell Wu,
  Sergey Edunov, Danqi Chen, and Wen-tau Yih. 2020.
\newblock Dense passage retrieval for open-domain question answering.
\newblock \emph{arXiv preprint arXiv:2004.04906}.

\bibitem[{Kwiatkowski et~al.(2019)Kwiatkowski, Palomaki, Redfield, Collins,
  Parikh, Alberti, Epstein, Polosukhin, Devlin, Lee
  et~al.}]{kwiatkowski2019natural}
Tom Kwiatkowski, Jennimaria Palomaki, Olivia Redfield, Michael Collins, Ankur
  Parikh, Chris Alberti, Danielle Epstein, Illia Polosukhin, Jacob Devlin,
  Kenton Lee, et~al. 2019.
\newblock Natural questions: a benchmark for question answering research.
\newblock \emph{Transactions of the Association for Computational Linguistics},
  7:453--466.

\bibitem[{Lee et~al.(2019)Lee, Chang, and Toutanova}]{lee2019latent}
Kenton Lee, Ming-Wei Chang, and Kristina Toutanova. 2019.
\newblock Latent retrieval for weakly supervised open domain question
  answering.
\newblock \emph{arXiv preprint arXiv:1906.00300}.

\bibitem[{Li and Liang(2021)}]{li2021prefix}
Xiang~Lisa Li and Percy Liang. 2021.
\newblock Prefix-tuning: Optimizing continuous prompts for generation.
\newblock \emph{arXiv preprint arXiv:2101.00190}.

\bibitem[{Liu et~al.(2021{\natexlab{a}})Liu, Yuan, Fu, Jiang, Hayashi, and
  Neubig}]{liu2021pre}
Pengfei Liu, Weizhe Yuan, Jinlan Fu, Zhengbao Jiang, Hiroaki Hayashi, and
  Graham Neubig. 2021{\natexlab{a}}.
\newblock Pre-train, prompt, and predict: A systematic survey of prompting
  methods in natural language processing.
\newblock \emph{arXiv preprint arXiv:2107.13586}.

\bibitem[{Liu et~al.(2021{\natexlab{b}})Liu, Ji, Fu, Du, Yang, and
  Tang}]{liu2021p}
Xiao Liu, Kaixuan Ji, Yicheng Fu, Zhengxiao Du, Zhilin Yang, and Jie Tang.
  2021{\natexlab{b}}.
\newblock P-tuning v2: Prompt tuning can be comparable to fine-tuning
  universally across scales and tasks.
\newblock \emph{arXiv preprint arXiv:2110.07602}.

\bibitem[{Liu et~al.(2019)Liu, Ott, Goyal, Du, Joshi, Chen, Levy, Lewis,
  Zettlemoyer, and Stoyanov}]{liu2019roberta}
Yinhan Liu, Myle Ott, Naman Goyal, Jingfei Du, Mandar Joshi, Danqi Chen, Omer
  Levy, Mike Lewis, Luke Zettlemoyer, and Veselin Stoyanov. 2019.
\newblock Roberta: A robustly optimized bert pretraining approach.
\newblock \emph{arXiv preprint arXiv:1907.11692}.

\bibitem[{Luan et~al.(2020)Luan, Eisenstein, Toutanova, and
  Collins}]{luan2020sparse}
Yi~Luan, Jacob Eisenstein, Kristina Toutanova, and Michael Collins. 2020.
\newblock Sparse, dense, and attentional representations for text retrieval.
\newblock \emph{arXiv preprint arXiv:2005.00181}.

\bibitem[{Mao et~al.(2020)Mao, He, Liu, Shen, Gao, Han, and
  Chen}]{mao2020generation}
Yuning Mao, Pengcheng He, Xiaodong Liu, Yelong Shen, Jianfeng Gao, Jiawei Han,
  and Weizhu Chen. 2020.
\newblock Generation-augmented retrieval for open-domain question answering.
\newblock \emph{arXiv preprint arXiv:2009.08553}.

\bibitem[{Ni et~al.(2021)Ni, Qu, Lu, Dai, {\'A}brego, Ma, Zhao, Luan, Hall,
  Chang et~al.}]{ni2021large}
Jianmo Ni, Chen Qu, Jing Lu, Zhuyun Dai, Gustavo~Hern{\'a}ndez {\'A}brego,
  Ji~Ma, Vincent~Y Zhao, Yi~Luan, Keith~B Hall, Ming-Wei Chang, et~al. 2021.
\newblock Large dual encoders are generalizable retrievers.
\newblock \emph{arXiv preprint arXiv:2112.07899}.

\bibitem[{Nogueira et~al.(2019)Nogueira, Lin, and
  Epistemic}]{nogueira2019doc2query}
Rodrigo Nogueira, Jimmy Lin, and AI~Epistemic. 2019.
\newblock From doc2query to doctttttquery.
\newblock \emph{Online preprint}.

\bibitem[{O{\u{g}}uz et~al.(2021)O{\u{g}}uz, Lakhotia, Gupta, Lewis, Karpukhin,
  Piktus, Chen, Riedel, Yih, Gupta et~al.}]{ouguz2021domain}
Barlas O{\u{g}}uz, Kushal Lakhotia, Anchit Gupta, Patrick Lewis, Vladimir
  Karpukhin, Aleksandra Piktus, Xilun Chen, Sebastian Riedel, Wen-tau Yih,
  Sonal Gupta, et~al. 2021.
\newblock Domain-matched pre-training tasks for dense retrieval.
\newblock \emph{arXiv preprint arXiv:2107.13602}.

\bibitem[{Qu et~al.(2021)Qu, Ding, Liu, Liu, Ren, Zhao, Dong, Wu, and
  Wang}]{qu2021rocketqa}
Yingqi Qu, Yuchen Ding, Jing Liu, Kai Liu, Ruiyang Ren, Wayne~Xin Zhao, Daxiang
  Dong, Hua Wu, and Haifeng Wang. 2021.
\newblock Rocketqa: An optimized training approach to dense passage retrieval
  for open-domain question answering.
\newblock In \emph{Proceedings of the 2021 Conference of the North American
  Chapter of the Association for Computational Linguistics: Human Language
  Technologies}, pages 5835--5847.

\bibitem[{Raffel et~al.(2019)Raffel, Shazeer, Roberts, Lee, Narang, Matena,
  Zhou, Li, and Liu}]{raffel2019exploring}
Colin Raffel, Noam Shazeer, Adam Roberts, Katherine Lee, Sharan Narang, Michael
  Matena, Yanqi Zhou, Wei Li, and Peter~J Liu. 2019.
\newblock Exploring the limits of transfer learning with a unified text-to-text
  transformer.
\newblock \emph{arXiv preprint arXiv:1910.10683}.

\bibitem[{Ren et~al.(2021)Ren, Qu, Liu, Zhao, She, Wu, Wang, and
  Wen}]{ren2021rocketqav2}
Ruiyang Ren, Yingqi Qu, Jing Liu, Wayne~Xin Zhao, Qiaoqiao She, Hua Wu, Haifeng
  Wang, and Ji-Rong Wen. 2021.
\newblock Rocketqav2: A joint training method for dense passage retrieval and
  passage re-ranking.
\newblock \emph{arXiv preprint arXiv:2110.07367}.

\bibitem[{Schick and Sch{\"u}tze(2020)}]{schick2020s}
Timo Schick and Hinrich Sch{\"u}tze. 2020.
\newblock It's not just size that matters: Small language models are also
  few-shot learners.
\newblock \emph{arXiv preprint arXiv:2009.07118}.

\bibitem[{Sun et~al.(2019)Sun, Wang, Li, Feng, Chen, Zhang, Tian, Zhu, Tian,
  and Wu}]{sun2019ernie}
Yu~Sun, Shuohuan Wang, Yukun Li, Shikun Feng, Xuyi Chen, Han Zhang, Xin Tian,
  Danxiang Zhu, Hao Tian, and Hua Wu. 2019.
\newblock Ernie: Enhanced representation through knowledge integration.
\newblock \emph{arXiv preprint arXiv:1904.09223}.

\bibitem[{Wang and Isola(2020)}]{wang2020understanding}
Tongzhou Wang and Phillip Isola. 2020.
\newblock Understanding contrastive representation learning through alignment
  and uniformity on the hypersphere.
\newblock In \emph{International Conference on Machine Learning}, pages
  9929--9939. PMLR.

\bibitem[{Xiong et~al.(2020)Xiong, Xiong, Li, Tang, Liu, Bennett, Ahmed, and
  Overwijk}]{xiong2020approximate}
Lee Xiong, Chenyan Xiong, Ye~Li, Kwok-Fung Tang, Jialin Liu, Paul Bennett,
  Junaid Ahmed, and Arnold Overwijk. 2020.
\newblock Approximate nearest neighbor negative contrastive learning for dense
  text retrieval.
\newblock \emph{arXiv preprint arXiv:2007.00808}.

\end{thebibliography}
\bibliographystyle{acl_natbib}

\appendix

\end{document}